\documentclass[Afour,sageh,times]{sagej}
\usepackage{moreverb,url}

\usepackage[ruled, linesnumbered]{algorithm2e}
\usepackage{subcaption}
\usepackage{url}
\usepackage{flushend}
\usepackage[latin1]{inputenc}
\usepackage{colortbl}
\usepackage{soul}
\usepackage{multirow}
\usepackage{multicol}
\usepackage{makecell}
\usepackage{booktabs}
\usepackage{pifont}
\usepackage{color}
\usepackage{alltt}
\usepackage{dsfont}
\usepackage[hidelinks]{hyperref}
\usepackage{enumerate}
\usepackage{siunitx}
\usepackage{breakurl}
\usepackage{epstopdf}
\usepackage{pbox}
\usepackage{algpseudocode}
\usepackage{array}
\usepackage{amssymb}
\usepackage{textcomp}
\usepackage{stfloats}

\newcommand\BibTeX{{\rmfamily B\kern-.05em \textsc{i\kern-.025em b}\kern-.08em T\kern-.1667em\lower.7ex\hbox{E}\kern-.125emX}}
\def\volumeyear{2022}
\begin{document}
\runninghead{Yin \MakeLowercase{\textit{et al.}}: \textbf{ALITA: A Large-scale Place Recognition Dataset for Long-term Autonomy.}}
\title{ALITA: A Large-scale Place Recognition Dataset for Long-term Autonomy}
\author{Peng Yin\affilnum{1}, Shiqi Zhao\affilnum{2,+}, Ruohai Ge\affilnum{1,+}, Ivan Cisneros\affilnum{1}, Ji Zhang\affilnum{1}, Howie Choset\affilnum{1} and Sebastian Scherer\affilnum{1}}

\affiliation{\affilnum{1}Robotics Institute, Carnegie Mellon University, USA.\\
\affilnum{2}University of California San Diego, USA. \\
\textsuperscript{+}Authors Shiqi Zhao and Ruohai Ge contributed equally to this paper.
}
\corrauth{Peng Yin, Robotics Institute, Carnegie Mellon University, Pittsburgh, PA 15213, USA.}
\email{pyin2@andrew.cmu.edu}

\begin{abstract}
For long-term autonomy, most place recognition methods are mainly evaluated on simplified scenarios or simulated datasets, which cannot provide solid evidence to evaluate the readiness for current Simultaneous Localization and Mapping (SLAM).
This paper presents a long-term place recognition dataset for use in mobile localization under large-scale dynamic environments.
This dataset includes a campus-scale track and a city-scale track.
The campus track focuses on the long-term property and is recorded with a LiDAR device and an omnidirectional camera on $10$ trajectories. 
Each trajectory is repeatedly recorded $8$ times under variant illumination conditions.
The city track focuses on the large-scale property and is recorded only with the LiDAR device on $120$km trajectory, which contains open streets, residential areas, natural terrains, etc.
They include $200$ hours of raw data of all kinds of scenarios within urban environments.
The ground truth position for both tracks is provided on each trajectory, obtained from the Global Position System with an additional General ICP-based point cloud refinement.
To simplify the evaluation procedure, we also provide the Python-API with a set of place recognition metrics proposed to quickly load our dataset and evaluate the recognition performance against different methods.
This dataset targets finding methods with high place recognition accuracy and robustness and providing real robotic systems with long-term autonomy.
We provide both the dataset and tools at
\href{https://github.com/MetaSLAM/ALITA}{https://github.com/MetaSLAM/ALITA}.
\end{abstract}


\maketitle

\begingroup
\let\clearpage\relax
\section{Introduction}
\label{sec:intro}
Place recognition or loop closure detection (LCD) is one of the most fundamental tasks in simultaneous localization and mapping (SLAM) and is also a key factor for long-term autonomy. 
In real-world environments, nevertheless, place recognition has been studied for decades~\cite{VPR:SURVEY}, reliable long-term and large-scale localization is still an unsolved problem. 
Recent years, noticeable developments in autonomous driving and last-mile delivery along with the increased demand for long-term, large-scale, and repeated (2LR) localization have been witnessed. 
Unlike the short-term SLAM tasks, the 2LR localization includes both spatial and temporary differences, and the existing datasets are either too complicated or too simple for evaluation methods. 
To assess the performance in localization tasks, most new recognition methods must be evaluated on previous exiting datasets, even with unique place feature encoding ability. 

With recent developments in computer vision, new learning- and non-learning-based, vision- and LiDAR-based place recognition methods have been proposed to improve the recognition performance under viewpoint and appearance differences.
All methods have their pros and cons according to the different environmental conditions.
Transitional non-learning based place recognition methods, such FABMAP~\cite{VPR:FABMAP}, CoHoG~\cite{cohog} and CALC~\cite{calc} for visual inputs, or ScanContext~\cite{scan_context}, M2DP~\cite{Feature3D:M2DP} and 3DSIFT~\cite{FEATURE3D:SIFT3D} for LiDAR inputs, have been well studied in the recent years but require careful parameter tuning.
In contrast, learning-based place recognition methods, such as NetVLAD~\cite{pr:netvlad}, NetVLAD~\cite{VPR:FABMAP}, PointNetVLAD~\cite{PR:pointnetvlad}, OverlapNet~\cite{PR:overlapnet} have shown improved place localization performance under complicate 3D/2D environment, such as the well-known \textit{KITTI} and \textit{NCLT} datasets.
    
    \begin{figure*}[t]
        \centering
        \includegraphics[width=\linewidth]{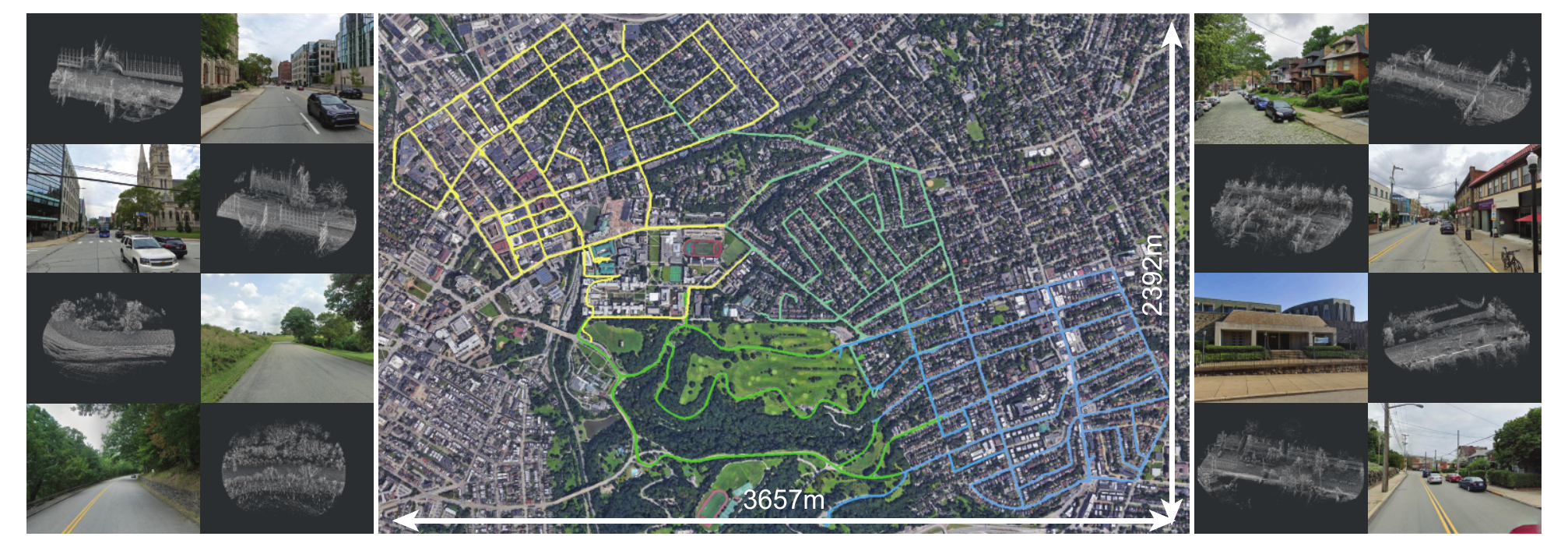}
        \caption{\textbf{ALITA Dataset(Urban).}
        The \textit{Urban} dataset includes four zones (colored in yellow, green, cyan, and blue) covering the downtown, residential, suburban, and commercial areas of Pittsburgh, Pennsylvania.
        }
        \label{fig:dataset_urban}
    \end{figure*}
    
    \begin{figure*}[t]
        \centering
        \includegraphics[width=\linewidth]{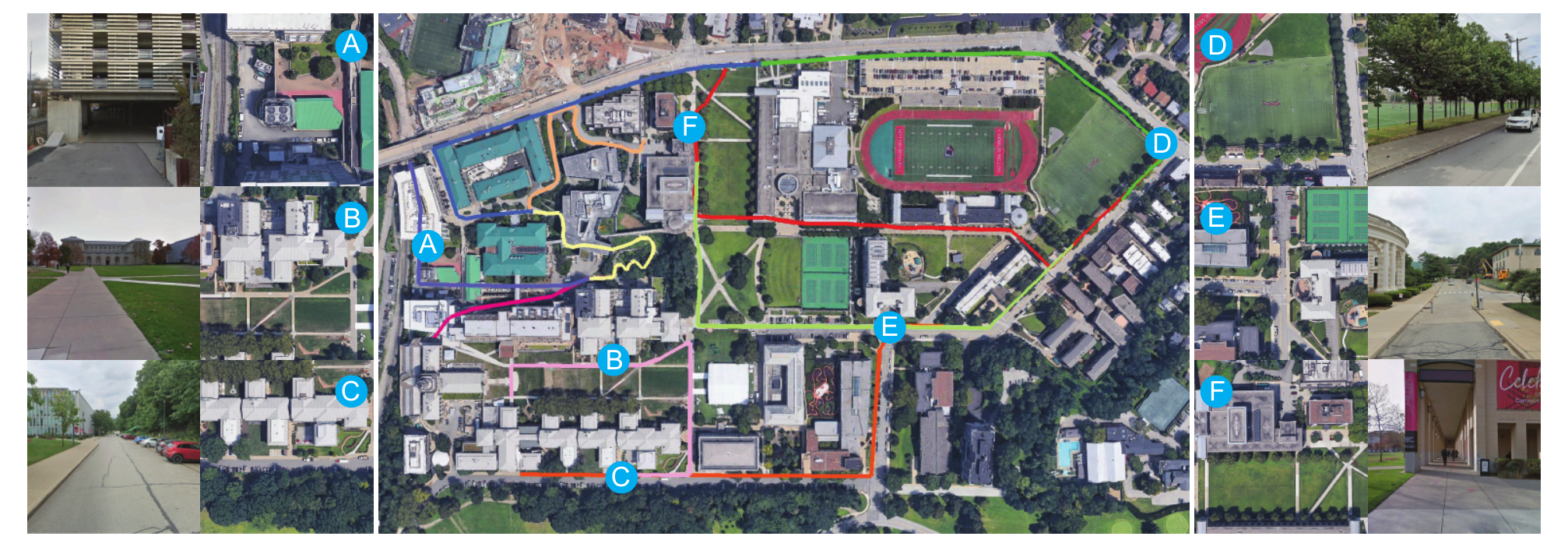}
        \caption{\textbf{ALITA Dataset(Campus).}
        The \textit{Campus} dataset covers the main campus area of Carnegie Mellon University and contains diverse 3D scenes including the buildings, parking lot, corridors, courts, and parks. 
        }
        \label{fig:dataset_campus}
    \end{figure*}

But currently, limited datasets can hardly evaluate the accuracy, robustness, and generalization ability of current methods under the (1) large-scale, (2) long-term, and (3) changing perspective environments.
Collecting a dataset that could cover the above three properties for real mobile platforms is a complicated task; the collection platform is hard and expensive to design, and the procedure for long-term and large-scale recording requires long-time preparation. 
Many existing methods are mainly evaluated on specific scenarios, and new researchers can hardly judge the performance in real applications.
Even the most recent place recognition approaches, LPDNet~\cite{pr:lpdnet} and OverlapNet~\cite{PR:overlapnet} are mainly claimed state-of-the-art performance over limited test datasets, and few methods have a reasonable number of real-world evaluations.

\begin{table*} [htbp]
    \caption{\textbf{Comparison of different map merging approaches.} 
    Total length is the multiplication of geographical coverage with the number of traversing.
    Temporal Diversity includes seasonal changes and day-night changes.
    "-" means not applicable.
    }
    \centering
    \begin{tabular}{|c|c|c|c|c|c|}
    \hline
    \textbf{Method}
    & \textbf{Scenarios}
    & \makecell[c]{\textbf{Total} \\ \textbf{Length}}
    & \makecell[c]{\textbf{Temporal} \\ \textbf{Diversity}}
    & \makecell[c]{\textbf{Viewpoint} \\ \textbf{Diversity}}
    & \makecell[c]{\textbf{Structural} \\ \textbf{Diversity}}
    \\
    \hline
    Freiburg~\cite{DATASET:Freiburg} & Campus & $\sim 0.7$ & One time & $-$ & $\star$\\
    \hline
    Ford Campus~\cite{DATASET:FordCampus} & Campus & $-$ & One time & $-$ & $\star$\\
    \hline
    KITTI~\cite{DATASET:KITTI} & Urban & $\sim 39.2$ & One time & $\star$ & $\star\star$\\
    \hline
    NCLT~\cite{DATASET:NCTL} & Campus & $\sim 148.5$ & Season & $\star$ & $\star$\\
    \hline
    Oxford RobotCar~\cite{DATASET:Oxford} & Urban & $\sim 1000$ & Day-Night & $\star$ & $\star$\\
    \hline
    MulRan~\cite{DATASET:mulran} & Campus \& Urban & $\sim 123.9$ & Multi times & $\star\star\star$ & $\star\star\star$\\
    \hline
    KITTI360~\cite{DATASET:KITTI360} & Urban & $\sim 1000$ & Day-Night & $\star\star$ & $\star\star$\\
    \hline
    ALITA(Urban)(ours)  & Urban & $\sim 120$ & One time & $\star\star$ & $\star\star\star$\\
    \hline
    ALITA(Campus)(ours) & Campus & $\sim 50$ & Day-Night & $\star\star\star$ & $\star$\\
    \hline
    \end{tabular}
\label{tab:CorrelationAveraging}
\end{table*}

This paper presents ALITA, a dataset set for long-term place recognition in large-scale environments.
Our datasets contain two tracks: (1) \textit{Urban} dataset, which records LiDAR data inputs in a city-scale urban-like area for $50$ segments and $120km$ trajectory in total.
And (2) \textit{Campus} dataset, recorded under a campus-scale environment, where we gathered the omnidirectional visual inputs and LiDAR inputs on $10$ different trajectories for $8$ repeated times, under different illuminations and viewpoints; this dataset targets long-term localization challenge.
Figure.~\ref{fig:dataset_urban} and Figure.~\ref{fig:dataset_campus} gives a better visualization of its scale, and Table.~\ref{tab:CorrelationAveraging} shows the comparison of different datasets.
Most datasets are targeted at short-term, fixed conditions or viewpoints place recognition tasks, so it is hard to evaluate the localization performance in real-world long-term, large-scale applications.
Compared to existing datasets, our \textit{Urban} dataset covers variant 3D scenarios for comprehensive 3D place recognition evaluation and multi-session SLAM~\cite{van2018collaborative,kimera_multi}.
And our \textit{Campus} dataset repeatedly covers diverse campus areas with dynamic objects, illumination, and viewpoint differences, which is suitable to evaluate long-term re-localization or incremental learning ability.

Both \textit{Urban} and \textit{Campus} datasets provide the ground truth for the exact place recognition to help evaluate different methods.
The \textit{Urban} dataset has been used in the IEEE ICRA 2022 \href{https://www.aicrowd.com/challenges/icra2022-general-place-recognition-city-scale-ugv-localization}{\textit{General Place Recognition Competition}} to benchmark the current new 3D place recognition approaches.
This paper describes the details of both datasets and provides the Python-API for the whole pipeline of data processing and localization evaluation in \href{https://github.com/MetaSLAM/ALITA}{https://github.com/MetaSLAM/ALITA}.

\section{Related works}
\label{sec:related_works}

    \subsection{Place Recognition}
    Based on the natural robustness of the LiDAR against illumination variant, loads of LiDAR-based place recognition methods have been developed.
    The success of PointNet~\cite{pointnet} makes it possible to extract the features from the point cloud directly and, therefore \cite{PR:pointnetvlad} convert the local features utilizing PointNet into a global descriptor via a NetVLAD layer~\cite{pr:netvlad}.
    However, point-based methods suffer from the fixed point number of input which cannot provide many structural details, and giant model size, which leads to low computation efficiency and high computation cost.
    \cite{PR:minkloc3d} use sparse 3D convolutions on a voxelized point cloud to extract local features.
    The introduction of sparse 3d convolution accelerates the process of local feature extraction and extracts the features within the points in each local neighborhood.
    Nevertheless, most point-based and voxel-based methods are sensitive to viewpoint differences common in real-life robot navigation.
    Some projection-based methods \cite{scan_context}, \cite{PR:overlapnet}, \cite{PR:seqspherevlad}, \cite{PR:overlaptransformer} claim viewpoint invariant.
    Our dataset aims to offer good test cases for models to test their robustness under different viewpoints and translation differences.
    In the Urban dataset, we provide python API to generate query and database frames based on the user's requirements.
    The Campus dataset provides forward and reversed loop closures with dynamic object disturbance.
    
    \subsection{Existing Datasets}
    
    
    Table.~\ref{tab:CorrelationAveraging} summaries a set of outdoor place recognition datasets.
    \textit{Freiburg} and \textit{Ford Campus} are both collected in campus environments, but the insufficient size makes it hard to train large networks.
    \textit{KITTI}~\cite{DATASET:KITTI} is recorded by a data collection device, including a 64-beam LiDAR (Velodyne HDL-64E) mounted on the car in Karlsruhe.
    There are slight viewpoint changes, and most of the revisit is in the same direction.
    \textit{KITTI360~}\cite{DATASET:KITTI360} contains more reverse revisit, but temporal diversity is also not considered.
    \textit{NCLT}~\cite{DATASET:NCTL} contains times changes in the Campus for 15 months, which includes seasonal changes and repeated sequences that can improve the model's robustness during training.
    \textit{Oxford}~\cite{DATASET:Oxford} is collected by a car-mounted device and covers the same $10km$ route twice every week for a year. 
    However, the scene these two datasets contain is limited to a single place, which lacks geographical diversity.
    Our proposed dataset contains around $170km$ long trajectories split into two sections Urban and Campus.
    \textit{Urban} dataset contains $120km$ of trajectories which provide abundant frames and structural diversity(building, forest, etc.).
    The $50km$ \textit{Campus} dataset provides deliberate revisit in both directions along with temporal and illumination diversity (day and night)
    
    Place recognition methods can not only contribute to the SLAM system to alleviate localization shift but be essential to large map merging systems.
    \cite{automerge} has proven the application of place recognition in offline and real-time map merging.
    \textit{MulRan}~\cite{DATASET:mulran} contains various types of revisit, such as reverse revisit and lane-level revisit.
    In addition, the dataset is recorded in different environments.
    Even though loads of datasets emphasize deliberately designed revisit, they omit the revisit between different trajectories, and the spatial scale is limited to relatively small areas.
    In ALITA, we provide trajectories with overlaps with adjacent ones both in \textit{Urban} and \textit{Campus}, which makes it possible for methods to evaluate either city-scale or campus-scale environments.
    
\section{The Platform}
\label{sec:platform}
Our data collection platform contains a Velodyne VLP-16 LiDAR scanner, Xsens MTI-300 inertial measurement units, and an Nvidia Jetson TX2 onboard computer.
For the extrinsic calibration between LiDAR and inertial measurement units, we follow the method mentioned in
\href{https://github.com/ethz-asl/lidar_align}{https://github.com/ethz-asl/lidar\_align}.
For the \textit{Urban} dataset, we mount our platform onto the top of a mobile vehicle and parallel with the GNSS position system to record the ground truth positions in the city-scale environments.
As shown in Fig.~\ref{fig:platform}, for the \textit{Campus} dataset, we mount the same platform onto the mobile rover robot, and with an additional THETA V omnidirectional camera on the top of the LiDAR device; this setup can provide time-synced LiDAR inputs and 360 visual inputs.

\begin{figure}[h]
	\centering
    \includegraphics[width=\linewidth]{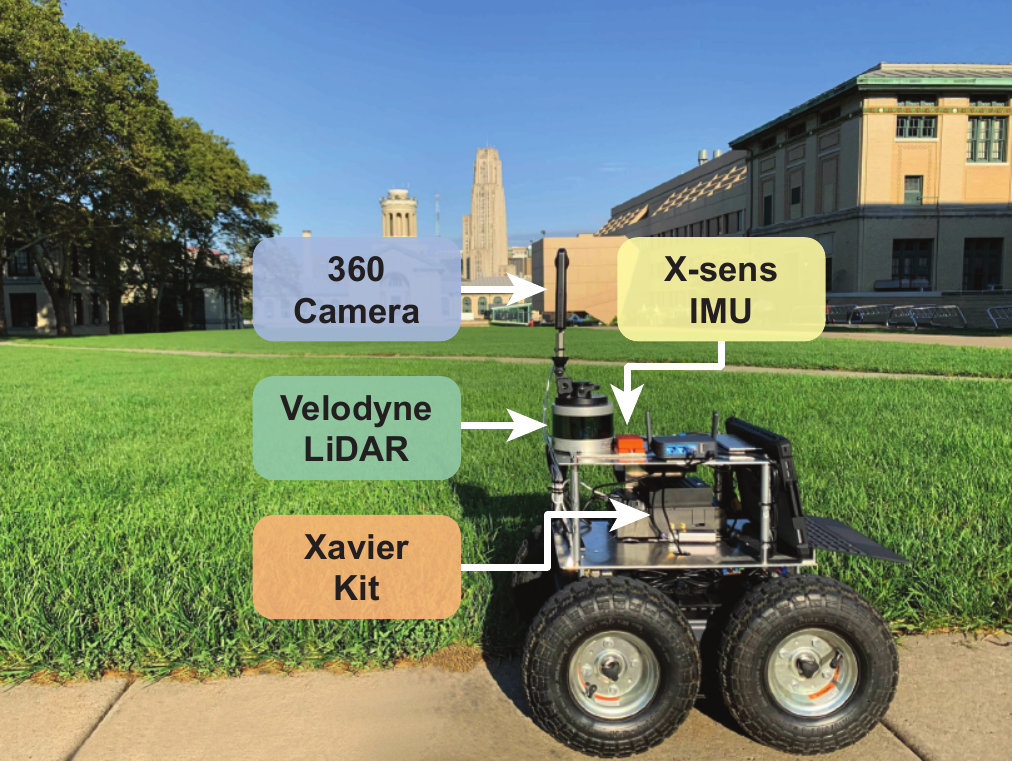}
	\caption{\textbf{Data-collection platform}}
	\label{fig:platform}
\end{figure}

\section{Dataset}
\label{sec:dataset}
\begin{figure*}[t]
    \centering
    \includegraphics[width=0.9\linewidth]{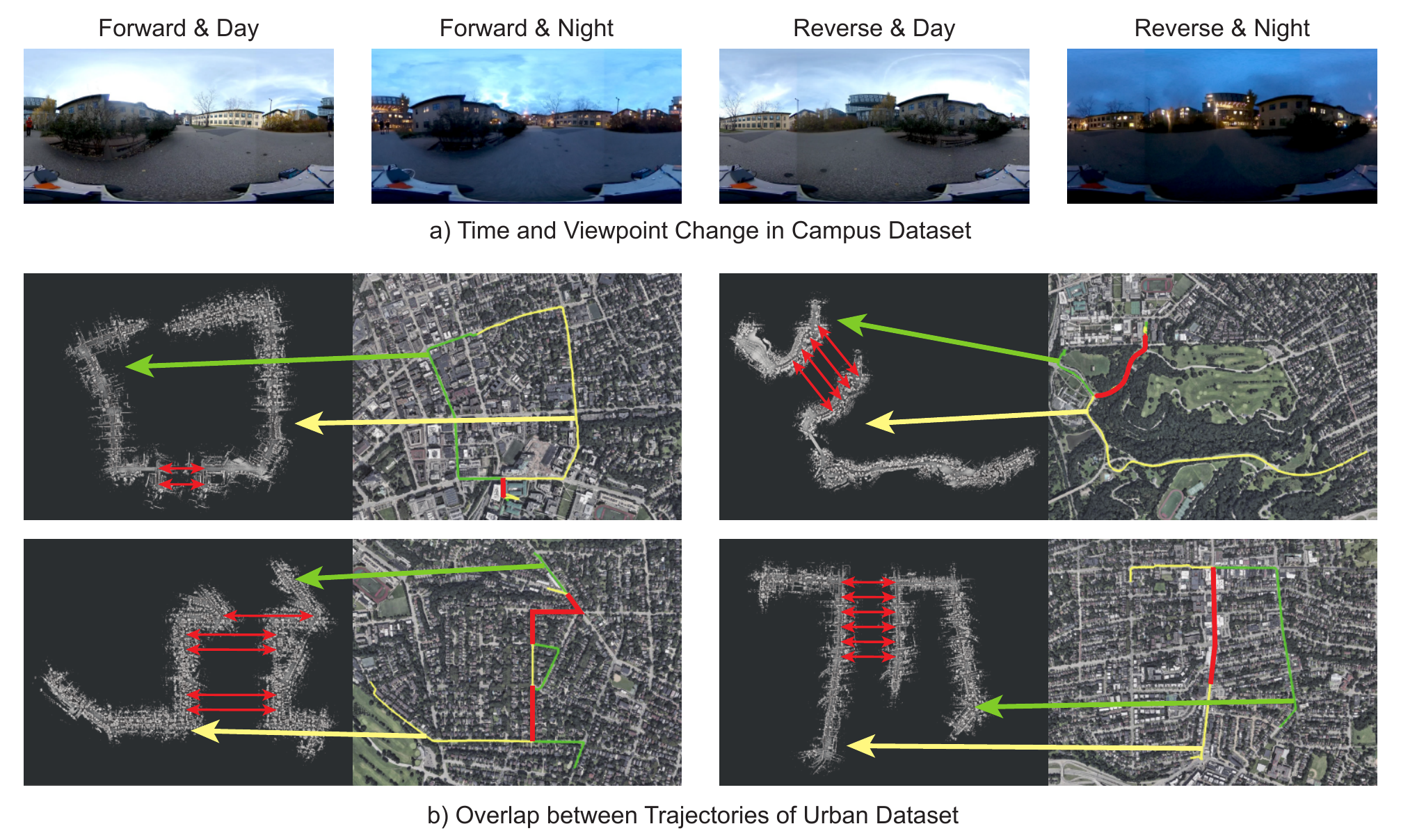}
    \caption{
    a) The Campus is equipped with an omnidirectional camera. As the data is recorded at different times of the day and from different viewpoints, the Campus can be utilized to test the re-localization ability under diverse illumination conditions.
    b) There are significant overlaps within trajectories of the Urban dataset.
    }
    \label{fig:dataset_details}
\end{figure*}

\subsection{Dataset Overview}
\label{sec:dataset_overview}
The present dataset contains two sub-datasets:

\textbf{Urban dataset} is composed of $50$ vehicle trajectories, as shown in Fig.~\ref{fig:dataset_urban}, covering $120km$ in the city of Pittsburgh, Pennsylvania;
The range of \textit{Urban} provides a sufficient quantity of data for extensive network training and high structural diversity, including commercial, residential, downtown, and suburban areas, for improving network robustness.
Especially, \textit{Urban} are also designed for the map-merging systems. 
As shown in Fig.~\ref{fig:dataset_details}, each trajectory is at least overlapped at one junction with the others, and there are $158$ overlaps in total within the dataset. 

\textbf{Campus dataset} consists of 10 trajectories collected within the campus area of Carnegie Mellon University(CMU), and the total length is around 36$km$.
Each trajectory is recorded eight times under different conditions(illumination, direction): as shown in Fig.~\ref{fig:dataset_details}, we have four types of combinations and recorded them two times each.
Even within a relatively small area, \textit{Campus} contains buildings, corridors, and crossroads, providing sufficient structural diversity for place recognition evaluation.
Moreover, omnidirectional pictures with position labels are provided, \textit{Campus} are also be utilized for visual place recognition evaluation under different viewpoint and illumination conditions.
Same as \textit{Urban}, there is a total of 9 overlaps.

\subsection{Data Description and Format}
Each trajectory of the Urban dataset consists of 3 types of data, described as follows:
\begin{itemize}
    \item[-] \textit{Global Maps}: Global maps are processed to contain the 3D structure of each trajectory, which is provided in Point Cloud Data (\texttt{PCD}) file format.
    We use self-developed LiDAR-Inertial odometry based on LOAM~\cite{LOAM:zhang2014loam} to generate global maps and process the maps with a VoxelGrid filter.
    \item[-] \textit{odometry}: We save the key poses generated by our SLAM algorithm as odometry information and provide them in (\texttt{TXT}) file format.
    The key poses are within the local coordinate of each trajectory, and the distance between adjacent poses is around $1m$.
    \item[-] \textit{Raw Data}: In order to offer convenience for map merging tasks, the raw data is provided in \texttt{rosbag} ROS package.
    Inside the raw data, two ROS topics \textit{/imu/data} and \textit{/velodyne\_packets} reveal the inertial measurement unit and LiDAR data.
    The frequency of two ROS topics is $200hz$ and $10hz$.
\end{itemize}

Each trajectory of the Campus dataset consists of 8 sequences, and each sequence includes four types of data:
\begin{itemize}
    \item[-] \textit{Global Maps} and \textit{Odometry} are the same with \textit{Urban}.
    \item[-] \textit{Unified Odometry}: We utilize interactive SLAM~\cite{interactiveslam} to find the geometric relations between the key poses of different sequences within the same trajectory and unify them into the same global coordinate.
    The data is provided in (\texttt{TXT}) file format.
    \item[-] \textit{Omnidirectional Pictures}: For each key pose, a corresponding omnidirectional picture with resolution of $1024 \times 512$ is provided in (\texttt{PNG}) file format.
\end{itemize}

\subsection{Using the Dataset}
The \texttt{PCD} files for global maps can be easily visualized using PCL~\cite{tools:PCL} or Open3D~\cite{tools:open3d} packages and the \texttt{bag} can be played back in the command line by \texttt{rosbag} ROS package.
In addition, we provide Python-API to access the data and generate training and validation sets based on global maps and corresponding odometry.
For \textit{Urban} dataset, we further provide highly personalized query and database frames generation to evaluate the performance of models on any translation and viewpoint differences.
As the \textit{Urban} dataset also can be used in large-scale map merging tasks, the ground truth of overlap between trajectories is provided in our API.
To offer convenience for models to compare with the state-of-art methods, we provide an online evaluation in AIcrowd to record each submission.

\begin{figure*}[t]
    \centering
    \includegraphics[width=\linewidth]{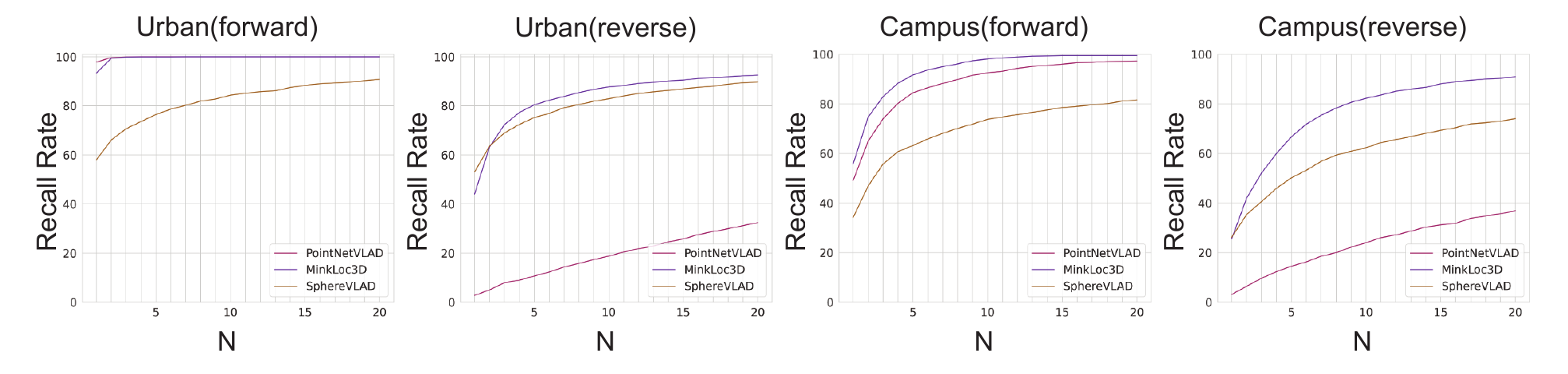}
    \caption{Methods of point-based, voxel-based and projection-based methods are selected to be tested on both datasets
    }
    \label{fig:experiment}
\end{figure*}
    
\section{Benchmark Experiments}
\label{sec:exp}

    \subsection{Models and Evaluation Methods}
    We select PointNetVLAD~\cite{PR:pointnetvlad}, MinkLoc3D~\cite{PR:minkloc3d} and SphereVLAD~\cite{PR:seqspherevlad} as our baseline models.
    Implementations in their official Github repository are utilized to train the networks from scratch on Urban dataset trajectory$(\texttt{01} \sim \texttt{10} \& \texttt{16} \sim \texttt{20})$ and the unified training data are available at \href{https://github.com/MetaSLAM/GPR_Competition}{\textit{General Place Recognition Competition}}.
    All models are tested in trajectory$(\texttt{21} \sim \texttt{41}$) of the Urban dataset for robustness under two combinations of translation and viewpoint differences denoted as forward and reverse.
    Queries frames are generated by uniformly sampling in the range of $[-3m \sim 3m]$ and $[-5^{\circ} \sim 5^{\circ}]$ based on database frames for forward and $[-3m \sim 3m]$ and $[175^{\circ} \sim 185^{\circ}]$ for reverse.
    Furthermore, trajectory$(\texttt{01} \sim \texttt{06})$ of \textit{Campus} are utilized to evaluate the generalization ability.
    We use Average Recall curve of top 20 candidates to show the performance of each method, and the Recall@1 is selected to compare each technique's retrieval and generalization ability. 
    A successful retrieval is defined as retrieving a point cloud within 5m for Pittsburgh and 3m for the Campus dataset.
    
    \subsection{Results}
    In the Urban dataset, PointNetVLAD outperforms the other methods at Recall@1 forward, and SphereVLAD exceeds the other methods at Recall@1 backward.
    Our Python-API can help researchers quickly analyze the recognition performance under variant viewpoints.
    Because the distance between adjacent database frames is only around 3m, the Campus dataset can be used to test the precision of re-localization.
    As shown in Fig.~\ref{fig:experiment}, all the models cannot achieve high recall@1 as in \textit{Urban} and MinkLoc3D outperforms other methods both in forward and backward.

\section{4. Summary and Future Work}
\label{sec:summary}
In this paper, we presented ALITA dataset which aims to long-term place recognition tasks in large-scale environments.
We believe that this dataset will be helpful in place recognition research in handling illumination and viewpoint changes and expect future LiDAR-Image(Omnidirectional) fusion-based robotics research.
Since the presented ALITA dataset provides abundant overlaps between trajectories, we also expect future usage for map merging systems.
We provided codes to help with using the dataset and evaluating new methods with it.

\section{5. Acknowledgment}
This research was supported by grants from NVIDIA and utilized NVIDIA SDKs (CUDA Toolkit, TensorRT, and Omniverse).
This research was also supported by ARL grant NO.W911QX20D0008.

\bibliographystyle{unsrtnat}
\bibliography{references}

\begin{thebibliography}{29}
\providecommand{\natexlab}[1]{#1}
\providecommand{\url}[1]{\texttt{#1}}
\expandafter\ifx\csname urlstyle\endcsname\relax
  \providecommand{\doi}[1]{doi: #1}\else
  \providecommand{\doi}{doi: \begingroup \urlstyle{rm}\Url}\fi

\bibitem[Lowry et~al.(2016)Lowry, Sünderhauf, Newman, Leonard, Cox, Corke, and
  Milford]{VPR:SURVEY}
Stephanie Lowry, Niko Sünderhauf, Paul Newman, John~J. Leonard, David Cox,
  Peter Corke, and Michael~J. Milford.
\newblock Visual place recognition: A survey.
\newblock \emph{IEEE Transactions on Robotics}, 32\penalty0 (1):\penalty0
  1--19, 2016.
\newblock \doi{10.1109/TRO.2015.2496823}.

\bibitem[Nowakowski et~al.(2017)Nowakowski, Joly, Dalibard, Garcia, and
  Moutarde]{VPR:FABMAP}
Mathieu Nowakowski, Cyril Joly, Sébastien Dalibard, Nicolas Garcia, and Fabien
  Moutarde.
\newblock Topological localization using wi-fi and vision merged into fabmap
  framework.
\newblock In \emph{2017 IEEE/RSJ International Conference on Intelligent Robots
  and Systems (IROS)}, pages 3339--3344, 2017.
\newblock \doi{10.1109/IROS.2017.8206171}.

\bibitem[Zaffar et~al.(2020)Zaffar, Ehsan, Milford, and McDonald-Maier]{cohog}
Mubariz Zaffar, Shoaib Ehsan, Michael Milford, and Klaus McDonald-Maier.
\newblock Cohog: A light-weight, compute-efficient, and training-free visual
  place recognition technique for changing environments.
\newblock \emph{IEEE Robotics and Automation Letters}, 5\penalty0 (2):\penalty0
  1835--1842, 2020.
\newblock \doi{10.1109/LRA.2020.2969917}.

\bibitem[Merrill and Huang(2018)]{calc}
Nate Merrill and Guoquan Huang.
\newblock Lightweight unsupervised deep loop closure.
\newblock \emph{CoRR}, abs/1805.07703, 2018.
\newblock URL \url{http://arxiv.org/abs/1805.07703}.

\bibitem[Kim and Kim(2018)]{scan_context}
Giseop Kim and Ayoung Kim.
\newblock Scan context: Egocentric spatial descriptor for place recognition
  within 3d point cloud map.
\newblock In \emph{2018 IEEE/RSJ International Conference on Intelligent Robots
  and Systems (IROS)}, pages 4802--4809, 2018.
\newblock \doi{10.1109/IROS.2018.8593953}.

\bibitem[He et~al.(2016)He, Wang, and Zhang]{Feature3D:M2DP}
Li~He, Xiaolong Wang, and Hong Zhang.
\newblock M2dp: A novel 3d point cloud descriptor and its application in loop
  closure detection.
\newblock In \emph{2016 IEEE/RSJ International Conference on Intelligent Robots
  and Systems (IROS)}, pages 231--237, 2016.
\newblock \doi{10.1109/IROS.2016.7759060}.

\bibitem[Mondal et~al.(2014)Mondal, Mukhopadhyay, Sural, and
  Bhattacharyya]{FEATURE3D:SIFT3D}
Prasenjit Mondal, Jayanta Mukhopadhyay, Shamik Sural, and Pinak~Pani
  Bhattacharyya.
\newblock 3d-sift feature based brain atlas generation: An application to early
  diagnosis of alzheimer's disease.
\newblock In \emph{2014 International Conference on Medical Imaging, m-Health
  and Emerging Communication Systems (MedCom)}, pages 342--347, 2014.
\newblock \doi{10.1109/MedCom.2014.7006030}.

\bibitem[Arandjelovic et~al.(2016)Arandjelovic, Gronat, Torii, Pajdla, and
  Sivic]{pr:netvlad}
Relja Arandjelovic, Petr Gronat, Akihiko Torii, Tomas Pajdla, and Josef Sivic.
\newblock Netvlad: Cnn architecture for weakly supervised place recognition.
\newblock In \emph{2016 IEEE Conference on Computer Vision and Pattern
  Recognition (CVPR)}, pages 5297--5307, 2016.
\newblock \doi{10.1109/CVPR.2016.572}.

\bibitem[Uy and Lee(2018)]{PR:pointnetvlad}
Mikaela~Angelina Uy and Gim~Hee Lee.
\newblock Pointnetvlad: Deep point cloud based retrieval for large-scale place
  recognition.
\newblock In \emph{2018 IEEE/CVF Conference on Computer Vision and Pattern
  Recognition}, pages 4470--4479, 2018.
\newblock \doi{10.1109/CVPR.2018.00470}.

\bibitem[Chen et~al.(2021)Chen, L{\"{a}}be, Milioto, R{\"{o}}hling, Vysotska,
  Haag, Behley, and Stachniss]{PR:overlapnet}
Xieyuanli Chen, Thomas L{\"{a}}be, Andres Milioto, Timo R{\"{o}}hling, Olga
  Vysotska, Alexandre Haag, Jens Behley, and Cyrill Stachniss.
\newblock Overlapnet: Loop closing for lidar-based {SLAM}.
\newblock \emph{CoRR}, abs/2105.11344, 2021.
\newblock URL \url{https://arxiv.org/abs/2105.11344}.

\bibitem[Liu et~al.(2019)Liu, Zhou, Suo, Yin, Chen, Wang, Li, and
  Liu]{pr:lpdnet}
Zhe Liu, Shunbo Zhou, Chuanzhe Suo, Peng Yin, Wen Chen, Hesheng Wang, Haoang
  Li, and Yunhui Liu.
\newblock Lpd-net: 3d point cloud learning for large-scale place recognition
  and environment analysis.
\newblock In \emph{2019 IEEE/CVF International Conference on Computer Vision
  (ICCV)}, pages 2831--2840, 2019.
\newblock \doi{10.1109/ICCV.2019.00292}.

\bibitem[Steder et~al.(2010)Steder, Grisetti, and Burgard]{DATASET:Freiburg}
Bastian Steder, Giorgio Grisetti, and Wolfram Burgard.
\newblock Robust place recognition for 3d range data based on point features.
\newblock In \emph{2010 IEEE International Conference on Robotics and
  Automation}, pages 1400--1405, 2010.
\newblock \doi{10.1109/ROBOT.2010.5509401}.

\bibitem[Pandey et~al.(2011)Pandey, McBride, and Eustice]{DATASET:FordCampus}
Gaurav Pandey, James~R McBride, and Ryan~M Eustice.
\newblock Ford campus vision and lidar data set.
\newblock \emph{The International Journal of Robotics Research}, 30\penalty0
  (13):\penalty0 1543--1552, 2011.
\newblock \doi{10.1177/0278364911400640}.
\newblock URL \url{https://doi.org/10.1177/0278364911400640}.

\bibitem[Geiger et~al.(2013)Geiger, Lenz, Stiller, and Urtasun]{DATASET:KITTI}
A~Geiger, P~Lenz, C~Stiller, and R~Urtasun.
\newblock Vision meets robotics: The kitti dataset.
\newblock \emph{The International Journal of Robotics Research}, 32\penalty0
  (11):\penalty0 1231--1237, 2013.
\newblock \doi{10.1177/0278364913491297}.
\newblock URL \url{https://doi.org/10.1177/0278364913491297}.

\bibitem[Carlevaris-Bianco et~al.(2016)Carlevaris-Bianco, Ushani, and
  Eustice]{DATASET:NCTL}
Nicholas Carlevaris-Bianco, Arash~K Ushani, and Ryan~M Eustice.
\newblock University of michigan north campus long-term vision and lidar
  dataset.
\newblock \emph{The International Journal of Robotics Research}, 35\penalty0
  (9):\penalty0 1023--1035, 2016.
\newblock \doi{10.1177/0278364915614638}.
\newblock URL \url{https://doi.org/10.1177/0278364915614638}.

\bibitem[Maddern et~al.(2017)Maddern, Pascoe, Linegar, and
  Newman]{DATASET:Oxford}
Will Maddern, Geoffrey Pascoe, Chris Linegar, and Paul Newman.
\newblock 1 year, 1000 km: The oxford robotcar dataset.
\newblock \emph{The International Journal of Robotics Research}, 36\penalty0
  (1):\penalty0 3--15, 2017.
\newblock \doi{10.1177/0278364916679498}.
\newblock URL \url{https://doi.org/10.1177/0278364916679498}.

\bibitem[Kim et~al.(2020)Kim, Park, Cho, Jeong, and Kim]{DATASET:mulran}
Giseop Kim, Yeong~Sang Park, Younghun Cho, Jinyong Jeong, and Ayoung Kim.
\newblock Mulran: Multimodal range dataset for urban place recognition.
\newblock In \emph{2020 IEEE International Conference on Robotics and
  Automation (ICRA)}, pages 6246--6253, 2020.
\newblock \doi{10.1109/ICRA40945.2020.9197298}.

\bibitem[Liao et~al.(2021)Liao, Xie, and Geiger]{DATASET:KITTI360}
Yiyi Liao, Jun Xie, and Andreas Geiger.
\newblock {KITTI-360:} {A} novel dataset and benchmarks for urban scene
  understanding in 2d and 3d.
\newblock \emph{CoRR}, abs/2109.13410, 2021.
\newblock URL \url{https://arxiv.org/abs/2109.13410}.

\bibitem[Van~Opdenbosch and Steinbach(2019)]{van2018collaborative}
Dominik Van~Opdenbosch and Eckehard Steinbach.
\newblock Collaborative visual slam using compressed feature exchange.
\newblock \emph{IEEE Robotics and Automation Letters}, 4\penalty0 (1):\penalty0
  57--64, 2019.
\newblock \doi{10.1109/LRA.2018.2878920}.

\bibitem[Tian et~al.(2022)Tian, Chang, Arias, Nieto-Granda, How, and
  Carlone]{kimera_multi}
Yulun Tian, Yun Chang, Fernando~Herrera Arias, Carlos Nieto-Granda, Jonathan~P.
  How, and Luca Carlone.
\newblock Kimera-multi: Robust, distributed, dense metric-semantic slam for
  multi-robot systems.
\newblock \emph{IEEE Transactions on Robotics}, pages 1--17, 2022.
\newblock \doi{10.1109/TRO.2021.3137751}.

\bibitem[Qi et~al.(2017)Qi, Su, Mo, and Guibas]{pointnet}
Charles~Ruizhongtai Qi, Hao Su, Kaichun Mo, and Leonidas~J. Guibas.
\newblock Pointnet: Deep learning on point sets for 3d classification and
  segmentation.
\newblock In \emph{2017 {IEEE} Conference on Computer Vision and Pattern
  Recognition, {CVPR} 2017, Honolulu, HI, USA, July 21-26, 2017}, pages 77--85.
  {IEEE} Computer Society, 2017.
\newblock \doi{10.1109/CVPR.2017.16}.
\newblock URL \url{https://doi.org/10.1109/CVPR.2017.16}.

\bibitem[Komorowski(2021)]{PR:minkloc3d}
Jacek Komorowski.
\newblock Minkloc3d: Point cloud based large-scale place recognition.
\newblock In \emph{2021 IEEE Winter Conference on Applications of Computer
  Vision (WACV)}, pages 1789--1798, 2021.
\newblock \doi{10.1109/WACV48630.2021.00183}.

\bibitem[Yin et~al.(2020)Yin, Wang, Egorov, Hou, Zhang, and
  Choset]{PR:seqspherevlad}
Peng Yin, Fuying Wang, Anton Egorov, Jiafan Hou, Ji~Zhang, and Howie Choset.
\newblock Seqspherevlad: Sequence matching enhanced orientation-invariant place
  recognition.
\newblock In \emph{2020 IEEE/RSJ International Conference on Intelligent Robots
  and Systems (IROS)}, pages 5024--5029, 2020.
\newblock \doi{10.1109/IROS45743.2020.9341727}.

\bibitem[Ma et~al.(2022)Ma, Zhang, Xu, Ai, Gu, and Chen]{PR:overlaptransformer}
Junyi Ma, Jun Zhang, Jintao Xu, Rui Ai, Weihao Gu, and Xieyuanli Chen.
\newblock Overlaptransformer: An efficient and yaw-angle-invariant transformer
  network for lidar-based place recognition.
\newblock \emph{IEEE Robotics and Automation Letters}, 7\penalty0 (3):\penalty0
  6958--6965, 2022.
\newblock \doi{10.1109/LRA.2022.3178797}.

\bibitem[Yin et~al.(2022)Yin, Lai, Zhao, Fu, Cisneros, Ge, Zhang, Choset, and
  Scherer]{automerge}
Peng Yin, Haowen Lai, Shiqi Zhao, Ruijie Fu, Ivan Cisneros, Ruohai Ge,
  Ji~Zhang, Howie Choset, and Sebastian Scherer.
\newblock Automerge: A framework for map assembling and smoothing in city-scale
  environments, 2022.
\newblock URL \url{https://arxiv.org/abs/2207.06965}.

\bibitem[Zhang and Singh(2014)]{LOAM:zhang2014loam}
Ji~Zhang and Sanjiv Singh.
\newblock {LOAM:} lidar odometry and mapping in real-time.
\newblock In Dieter Fox, Lydia~E. Kavraki, and Hanna Kurniawati, editors,
  \emph{Robotics: Science and Systems X, University of California, Berkeley,
  USA, July 12-16, 2014}, 2014.
\newblock \doi{10.15607/RSS.2014.X.007}.
\newblock URL \url{http://www.roboticsproceedings.org/rss10/p07.html}.

\bibitem[Koide et~al.(2021)Koide, Miura, Yokozuka, Oishi, and
  Banno]{interactiveslam}
Kenji Koide, Jun Miura, Masashi Yokozuka, Shuji Oishi, and Atsuhiko Banno.
\newblock Interactive 3d graph slam for map correction.
\newblock \emph{IEEE Robotics and Automation Letters}, 6\penalty0 (1):\penalty0
  40--47, 2021.
\newblock \doi{10.1109/LRA.2020.3028828}.

\bibitem[Rusu and Cousins(2011)]{tools:PCL}
Radu~Bogdan Rusu and Steve Cousins.
\newblock 3d is here: Point cloud library (pcl).
\newblock In \emph{2011 IEEE International Conference on Robotics and
  Automation}, pages 1--4, 2011.
\newblock \doi{10.1109/ICRA.2011.5980567}.

\bibitem[Zhou et~al.(2018)Zhou, Park, and Koltun]{tools:open3d}
Qian-Yi Zhou, Jaesik Park, and Vladlen Koltun.
\newblock Open3d: A modern library for 3d data processing, 2018.
\newblock URL \url{https://arxiv.org/abs/1801.09847}.

\end{thebibliography}
\endgroup

\end{document}